# Designs, motion mechanism, motion coordination, and communication of bionic robot fishes: a survey

Zhiwei Yu[1], Kai Li[1], Yu Ji[1], Simon X. Yang[2]

[1] College of Mechanical and Electrical Engineering, Nanjing University of Aeronautics and Astronautics, Nanjing 210000, China.
[2] School of Engineering, University of Guelph, 50 Stone Road East, Guelph, ON N1G 2W1, Canada.

## ABSTRACT

In the last few years, there have been many new developments and significant accomplishments in the research of bionic robot fishes. However, in terms of swimming performance, existing bionic robot fishes lag far behind fish, prompting researchers to constantly develop innovative designs of various bionic robot fishes. In this paper, the latest designs of robot fishes are presented in detail, distinguished by the propulsion mode. New robot fishes mainly include soft robot fishes and rigid-soft coupled robot fishes. The latest progress in the study of the swimming mechanism is analyzed on the basis of summarizing the main swimming theories of fish. The current state-of-the-art research in the new field of motion coordination and communication of multiple robot fishes is summarized. The general research trend in robot fishes is to utilize more efficient and robust methods to best mimic real fish while exhibiting superior swimming performance. The current challenges and potential future research directions are discussed. Various methods are needed to narrow the gap in swimming performance between robot fishes and fish. This paper is a first step to bring together roboticists and marine biologists interested in learning state-of-the-art research on bionic robot fishes.

## 1 Introduction

Propellers are frequently used as actuators in conventional underwater robots, and their propulsion efficiency is only 40%-50%. Furthermore, their shapes employ a non-bionic structure that cannot be integrated into the underwater environment, making close observation of underwater organisms difficult. Fish have undergone extensive natural selection and can swim with an efficiency of more than 90% [1]. Fish also have distinct advantages in terms of speed, maneuverability, and stealth [2-6]. For example, swordfish can reach a speed up to 30 m·s$^{-1}$ [3]. Bionic robot fishes, which treat fish as bionic objects, can effectively absorb these advantages to overcome the defects of traditional underwater robots and become more effective tools for ocean exploration.

The propulsion modes of fish are usually classified into two categories according to the body parts used for propulsion, namely body and/or caudal fin (BCF) propulsion and median and/or paired fin (MPF) propulsion [7,8]. It is worth noting that the median fin refers to the dorsal or anal fin, while the paired fin refers to the pectoral or pelvic fin. Taking tilapia as an example, the structure and position of each fin are shown in Figure 1. The BCF propulsion mode, in which the body and/or caudal fin acts as a propeller, is the most common in fish and first discovered by researchers. This propulsion mode has the advantages of high swimming speed and quick start performance, making it suitable for applications requiring high speed or instantaneous acceleration [9]. The median and/or paired fin acts as a propeller in the MPF

Designs, motion mechanism, motion coordination, and communication of bionic robot fishes: a survey propulsion mode. This propulsion mode has the advantages of high maneuverability, high propulsion efficiency, and good stability, making it suitable for applications requiring maneuvering to turn or long-term swimming, as well as scenes with rapid water flow [10]. After summarizing recent research results, we show that existing robot fishes already have the BCF and MPF combined (BCF-MPF) propulsion mode. This propulsion mode is based on the cooperation of the caudal and pectoral fins. With proper design, it is capable of balancing swimming speed and propulsion efficiency, which has a wider application than either individually. Furthermore, it is a promising research topic. The basic elements of the three propulsion modes are summarized in Table 1.

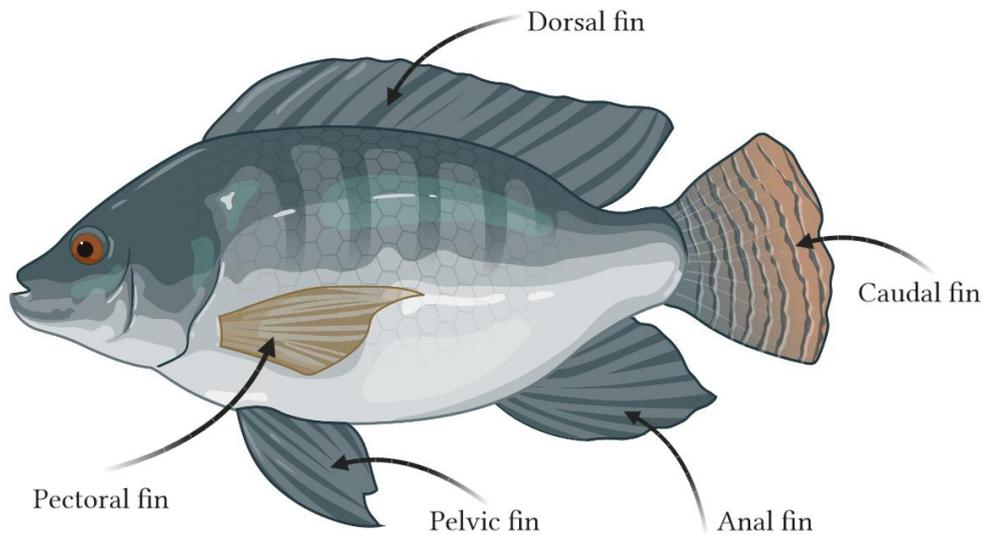

Figure 1: Types of fins in tilapia.

Table 1: Propulsion modes of robot fishes.

| Propulsion modes | Propellers | Strengths | Applications |
| --- | --- | --- | --- |
| BCF | Body and/or caudal fin | 1. High swimming speed<br>2. High quick start performance | 1. Requiring high speed<br>2. Requiring instantaneous acceleration |
| MPF | Median and/or paired fin | 1. High maneuverability<br>2. High propulsion efficiency<br>3. Good stability | 1. Requiring maneuvering to turn<br>2. Requiring long-term swimming<br>3. Rapid water flow |
| BCF and MPF | Cooperation of the caudal and pectoral fins | Balancing swimming speed and propulsion efficiency | Broader than either individually |

Some review papers focus on the motion control of robot fishes [11-14], while others focus on the design, fabrication, and propulsion methods of robot fishes [15-17]. There is also a review paper that focuses on the perception of robot fishes [18]. However, the majority of them were published more than five years ago. In these years, unprecedented attention has been paid to the study of bionic robot fishes. Related achievements have proliferated and enriched the research in the field of robot fishes. As a result, this paper provides a new survey on various major fields of robot fishes, addressing some gaps in related fields. Figure 2 depicts the paper's framework, which includes three objectives. Firstly, we provide a comprehensive survey of the most recent designs of robot fishes, as well as the most recent progress in the study of motion mechanism. A new field of study, namely motion coordination and communication of multiple robot fishes, is discussed. Second, based on the survey, the challenges of current research and potential future research directions are summarized. Three aspects are included: the gap between robot fishes and fish in terms of swimming performance, methods to study the swimming mechanism of robot fishes, and the motion coordination and communication of multiple robot fishes. Finally, a summary of the paper is provided.

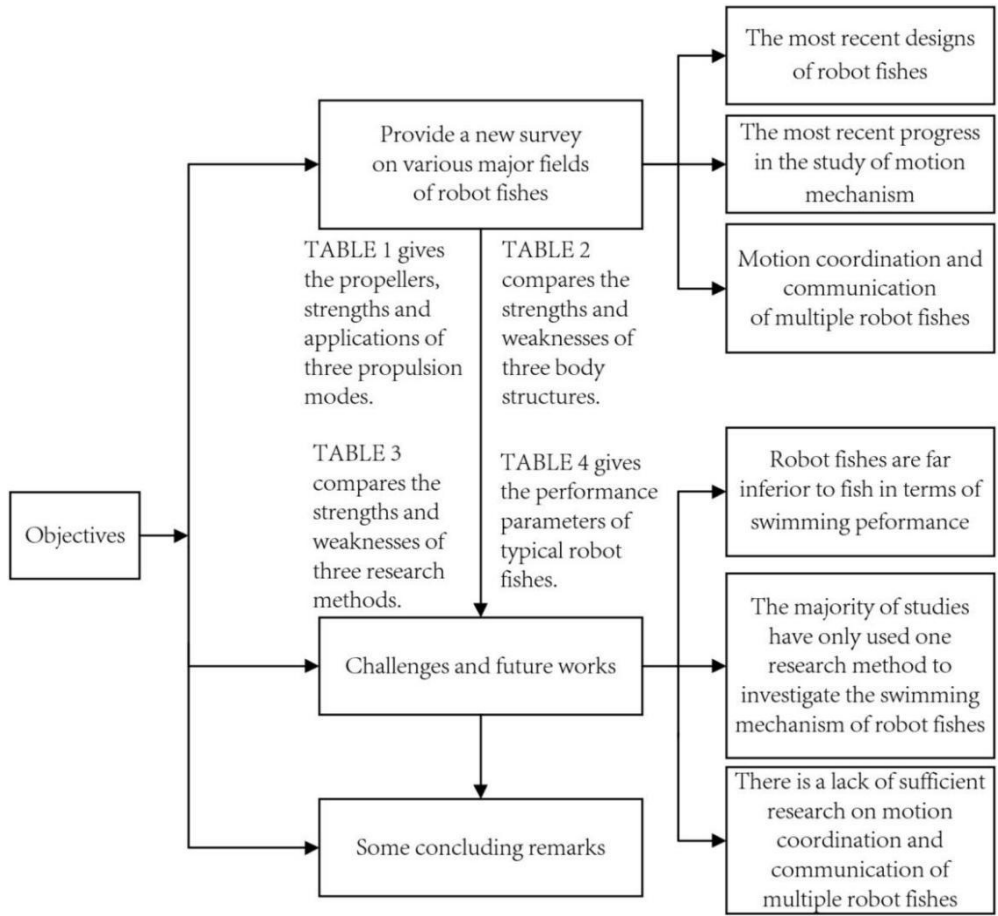

Figure 2: The review framework.

The rest of the article is organized as follows. Section 2 elaborates on the latest designs of robot fishes. Section 3 analyzes the motion mechanism of robot fishes. The motion coordination and communication of multiple robot fishes are discussed in Section 4. Section 5 provides a comprehensive discussion on the challenges and future works. Finally, in Section 6, some concluding remarks are made.

## 2 Designs of robot fishes

According to their body structure, robot fishes are classified into three types: rigid, soft, and rigid-soft coupled. The strengths and weaknesses of different body structures are summarized in Table 2. The rigid robot fish has high swimming speed, but its maneuverability is poor. In contrast, the soft robot fish has great maneuverability, but its swimming speed is low. The rigid-soft coupled robot fish lies between the two. Through reasonable design, it can achieve great maneuverability while generating high swimming speed. The rigid robot fish has received little attention in recent years. This is primarily due to the fact that the rigid structure of the rigid robot fish is far from the elastic skin and muscles of fish. As a result, we only discuss soft and rigid-soft coupled robot fishes in this paper.

Designs, motion mechanism, motion coordination, and communication of bionic robot fishes: a survey

Table 2: The body structures of robot fishes.

| Body structures | Strengths | Weaknesses |
|---|---|---|
| Rigid | High swimming speed | Poor maneuverability |
| Soft | Great maneuverability | Low swimming speed |
| Rigid–soft coupled | Achieving great maneuverability while generating high swimming speed with a reasonable design | |

## 2.1 Robot fishes in BCF propulsion mode
### 2.1.1 Soft robot fishes

This robot fish generally uses intelligent materials or other special devices to simulate the muscles of fish, expecting to significantly improve the swimming performance. The first attempt was made by Katzschmann *et al.* [19]. The robot fish SoFi was designed by them, which used a soft fluid actuator to simulate muscle tissue. The soft caudal fin had two lateral chambers symmetrical along the central axis. A gear pump drove fluid flow from one side of the chamber to another, causing the caudal fin to bend. SoFi successfully swam around aquatic life at depths of 0-18 m and effectively integrated into the marine environment. However, SoFi still has room for improvement, such as optimizing the geometry of the tail section. Dielectric elastomer actuators (DEAs), a type of smart material, are also used in robot fishes. Shintake *et al.* attached two DEAs to both sides of the robot fish's body, as shown in Figure 3A [20]. The DEAs were stretched by the same length so that the initial state of the robot fish was straight. The voltage was applied to each side of the body in turn, so that one side of the DEA was elongated while the other side was contracted. As a result, the robot fish's body oscillated from one side to another side, causing the caudal fin to oscillate. The maximum swimming speed of the robot fish was 0.25 BL·s$^{-1}$ at 0.75 hertz oscillation frequency. However, the authors needed to test the fish in different sizes and swimming types (e.g., turning) to figure out how much swimming ability the fish had. Liu *et al.* proposed using gas-driven units to simulate muscle fibers of fish and successfully designed the robot fish Flexi-Tuna [21]. As shown in Figure 3B, 14 drive units were symmetrically distributed on both sides of the robot fish's body. Then, alternating pressure was applied to the drive units to make the tail oscillate back and forth. According to the results, under the optimal frequency of 3.5 Hz, the maximum swing angle of Flexi-Tuna was 20° and the maximum thrust was 0.185 N. This research realized the application of artificial muscles in robot fishes and provided new ideas for the design of soft robot fishes. However, some optimizations, such as variable stiffness design of caudal fin, are still needed to achieve better swimming performance of robot fishes.

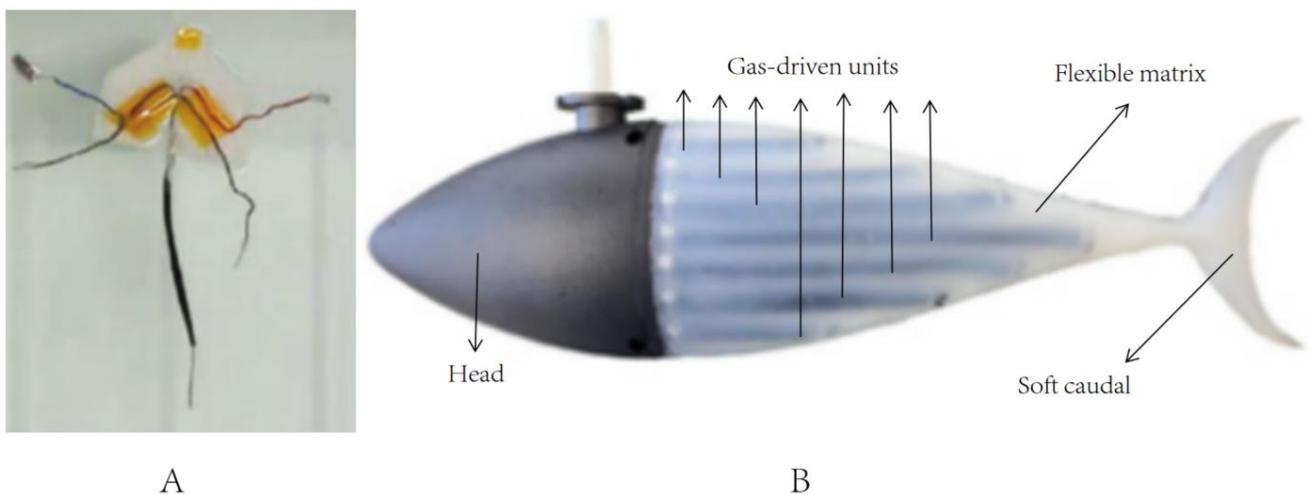

Figure 3: Soft robot fishes in BCF propulsion mode: (A) a robot fish with DEAs [20]; and (B) Flexi-Tuna [21]. BCF, body and/or caudal fin; DEAs, dielectric elastomer actuators.

### 2.1.2 Rigid–soft coupled robot fishes

In recent years, researchers have come up with some new ideas to improve the swimming performance of this type of robot fish.

The headshaking of robot fishes leads to an increase of water resistance, which in turn reduces their swimming speed. To address this issue, Liao *et al*. proposed using two caudal fins rather than a single caudal fin [22]. Caudal fins were mounted symmetrically on the tail of the robot fish, as shown in Figure 4A. They were designed to flap oppositely to offset lateral forces, which in turn prevented the headshaking. The robot fish had three motions: oscillatory motion, jet motion, and oscillatory and jet cooperative motion. A suitable motion type could be chosen based on the distance between two caudal fins. This indicated that the robot fish had great flexibility. According to the experimental results, the robot fish could reach the speed of 2.5 body lengths per second (BL·s$^{-1}$), demonstrating excellent swimming speed.

Researchers have made great progress in mimicking the body structure of fish. Coral *et al*. created a robot fish using actuators made of shape memory alloys (SMAs) [23]. As shown in Figure 4B, these actuators were bent into a continuous structure to resemble the fish backbone. Bio-inspired synthetic skin was used to mimic the skin of fish. Nevertheless, the authors only verified the feasibility of this scheme. Zhu *et al*. created Tunabot by mimicking the body structure of tuna and mackerel and discussed the influence of oscillation frequency in depth [24]. The robot fish had a streamlined shape with an elastic skin overlaid on the actuator system. Tunabot swam at a maximum tail-beat frequency of 15 hertz, reaching 4 BL·s$^{-1}$ according to experiments. Tunabot could swim 9.1 km if it swam at 0.4 m·s$^{-1}$ or 4.2 km if it swam at 1 m·s$^{-1}$ while powered by a 10 Wh battery pack. This highlighted the capabilities of high-frequency swimming. This

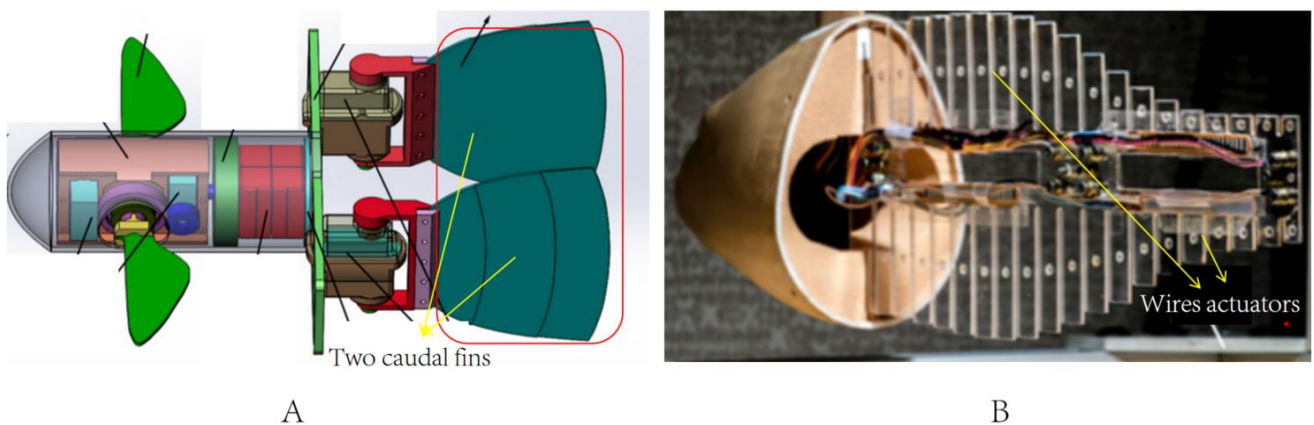

Figure 4: Rigid-soft coupled robot fishes in BCF propulsion mode: (A) a robot fish with two caudal fins [22]; and (B) wires actuators of the robot fish [23]. BCF, body and/or caudal fin.

Designs, motion mechanism, motion coordination, and communication of bionic robot fishes: a survey provided new ideas to improve the swimming performance of robot fishes. The variable stiffness design of the robot fish is also an imitation of fish. TenFiBot, a robot fish with variable stiffness, was designed by Chen and Jiang [25]. The whole structure of TenFiBot was a tandem structure with multiple variable-stiffness tensegrity joints (VSTJs). The preload of the springs on the VSTJs could be adjusted to change the stiffness distribution on the TenFiBot's body. Experiments demonstrated that the change of stiffness distribution directly affected the swimming performance (such as swimming speed) of the robot fish. By changing the stiffness distribution of the robot fish, its swimming performance could be greatly improved.

## 2.2 Robot fishes in MPF propulsion mode
### 2.2.1 Soft robot fishes

This robot fish tends to be designed with smart materials and is smaller in size. As MPF propulsion mode is adopted, it has greater maneuverability. Therefore, it is ideal for applications in special environments, such as fine pipes, deep sea, etc. Inspired by the hadal snail-fish, which lives at 8000 m water depth, Li *et al.* designed an untethered soft robot fish that could withstand extreme hydrostatic pressure [26]. The robot fish was driven by DEAs. The electronic components of the robot fish were decentralized on several smaller printed circuit boards, which could effectively reduce the shear stress between components. This ensured that the robot fish could withstand extreme water pressure. The robot fish successfully swam at a depth of 10,900 m in the Mariana Trench, showing great potential for application in deep-sea exploration.

### 2.2.2 Rigid–soft coupled robot fishes

A key condition to achieving high swimming performance is to adjust the distribution of soft and hard structures in robot fishes. As shown in Figure 5A, a robot fish with cartilages and soft tissues was designed by Yurugi *et al.* [27]. Experiments revealed that adding cartilages to the fins of the robot fish could improve swimming efficiency. The researchers also investigated the fish's swimming behavior. As shown in Figure 5B, Ma *et al.* designed a robot fish driven by the oscillating and twisting of the pectoral fins after studying the pectoral fin movement of the cownose ray [28]. The pectoral fins simultaneously realized oscillating motion and chordwise twisting motion. The maximum swimming speed of the robot fish was 0.94 $BL \cdot s^{-1}$, and the turning radius was nearly zero. This reflected the excellent turning performance and high swimming speed of the robot fish. These authors should conduct additional research into the effect of pectoral fin flexibility on swimming performance.

## 2.3 Robot fishes in BCF-MPF propulsion mode
### 2.3.1 Soft robot fishes

The caudal fin of this robot fish mainly serves a steering function, and the pectoral fins mainly provide propulsion. Zhang *et al.* first tried to build a robot fish [29]. The dielectric elastomers (DEs) were attached to the elastic frame, and the variable voltage was applied to drive the pectoral fins up and down to generate forward thrust. A steering electrical servo drove the caudal fin deflection angle for turning. Figure 6A depicts the position of the pectoral and caudal fins. Unfortunately, the authors did not test the performance of this robot fish. Li *et al.*, inspired by manta rays, designed a soft electronic robot fish driven by DEAs, as shown in Figure 6B [30]. The speed of the robot fish was 0.69 $BL \cdot s^{-1}$. It could use the surrounding water as an electric ground and swim for up to 3 h on a single charge. This thoroughly illustrated the robustness of this robot fish.

Designs, motion mechanism, motion coordination, and communication of bionic robot fishes: a survey

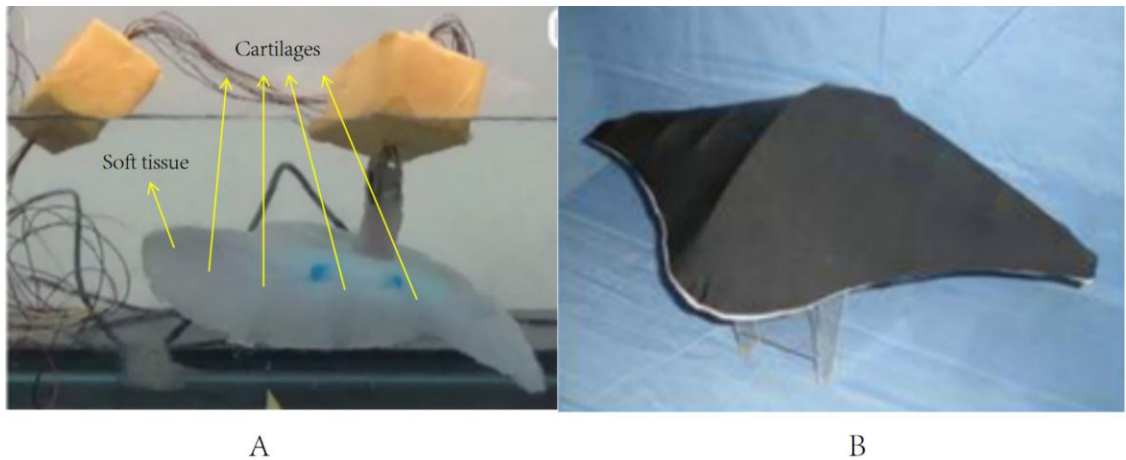

Figure 5: Rigid–soft coupled robot fishes in MPF propulsion mode: (A) a robot fish with soft tissue and cartilages. [27]; and (B) bionic cownose ray robot fish [28]. MPF, median and/or paired fin.

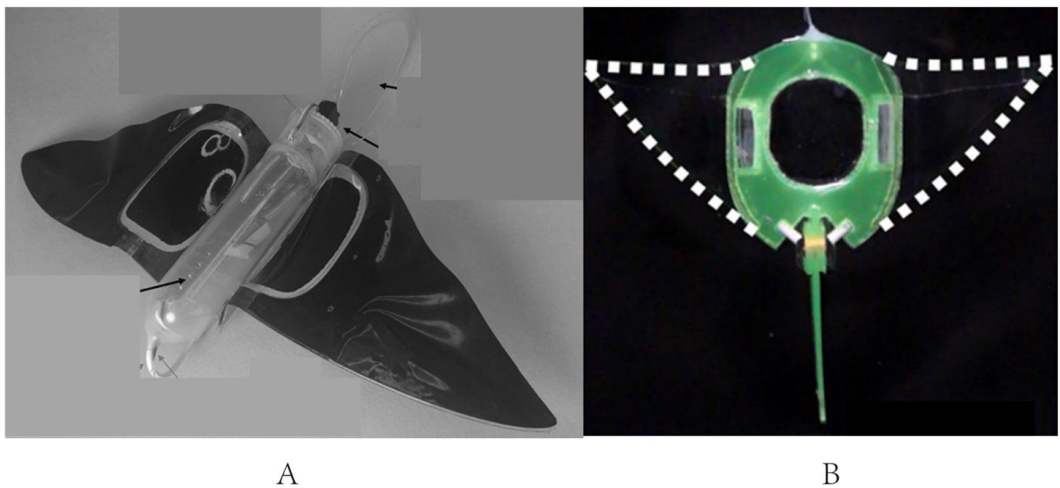

Figure 6: Soft robot fishes in BCF-MPF propulsion mode: (A) a soft robot fish [29]; and (B) a soft electronic fish [30]. BCF, body and/or caudal fin; MPF, median and/or paired fin.

### 2.3.2 Rigid–soft coupled robot fishes

The caudal fin of this robot fish is able to oscillate significantly and rapidly, allowing for high propulsion power. Simultaneously, the pectoral fins have multiple degrees of freedom, allowing for great maneuverability. As a result, the excellent swimming performance of these robot fishes has attracted the interest of many researchers. This was attempted by Li *et al.*, who created the robot fish shown in Figure 7 [31]. The caudal fin of the robot fish had three rigid joints, which ensured its high flexibility. The pectoral fins could perform rotary motion and forward–backward motion, and the two motions were completely independent. This robot fish could reach a turning speed of 0.6 radians per second (rad·s$^{-1}$) with the coordinated propulsion of the caudal and pectoral fins. This provided the robot fish with more turning options and higher maneuverability. Zhong *et al.* designed a new type of robot fish [32]. The caudal fin of the robot fish was driven by wires, which could be deformed along a chordwise direction or both chordwise and spanwise directions. Flapping and rowing motions were possible with the pectoral fins. The results show that, without using the pectoral fins, the turning radius of the robot fish was 0.6 BL; with the pectoral fins, the turning radius was reduced to 0.25 BL. This clearly had higher maneuverability. These experiments only tested the turning performance of these robot fishes and did not test their performance in other swimming types (e.g., straight swimming).

Designs, motion mechanism, motion coordination, and communication of bionic robot fishes: a survey

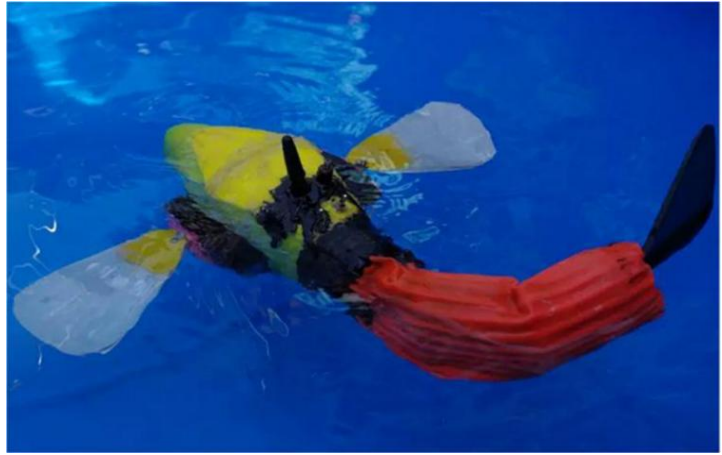

Figure 7: A rigid–soft coupled robot fish in BCF-MPF propulsion mode [31]. BCF, body and/or caudal fin; MPF, median and/or paired fin.

## 3 The motion mechanism of robot fishes

The study of the motion mechanism of robot fishes provides an in-depth understanding of the process by which robot fishes obtain thrust. The results of this study can be utilized to improve the designs of robot fishes. This enables robot fishes to achieve higher propulsion power and efficiency, bridging the swimming performance gap between robot fishes and fish.

Currently, there are three main research methods to study the swimming mechanism of robot fishes. The strengths and weaknesses of the three research methods are summarized in Table 3. The first method is theoretical analysis. In this method, the swimming equations of robot fishes are established by mathematical and physical models. The method is very adaptable, but it is mathematically challenging. Further, the difficulty lies in the need to establish equations that can be solved and correctly describe the complex swimming of robot fishes. The second method is experimental observation. This method uses particle image velocimetry (PIV) or other special equipment to observe robot fishes or fish. The conclusions of the research are highly accurate due to real-world observations, but they have poor universality due to the experimental setting's restrictions. The third method is numerical simulation, which uses computers to numerically solve existing models to predict the swimming characteristics of robot fishes. The method is low cost and accurate, but it cannot solve some complex swimming problems that lack a perfect mathematical model. We can see that each of the three research methods has strengths and weaknesses, and combining these methods can yield complementary benefits.

Table 3. Research methods to study the swimming mechanism of robot fishes.

| **Research methods** | **Strengths** | **Weaknesses** |
| --- | --- | --- |
| Theoretical analysis | Very adaptable | Mathematically challenging |
| Experimental observation | Highly accurate | Poor universality |
| Numerical simulation | 1. Low cost<br>2. Accurate | Solving a limited number of problems |

Designs, motion mechanism, motion coordination, and communication of bionic robot fishes: a survey

## 3.1 Theoretical analysis

The swimming of robot fishes mimics that of fish. A better understanding of the fish motion mechanism aids in the design of robot fishes. There are numerous theories about fish swimming, but only a few widely accepted ones are discussed here. In 1970, Lighthill proposed the "elongated-body theory" [33]. This theory only investigates the role of the fish's caudal cross-section in swimming, ignoring the effect of the caudal vortex. As a result, the swimming performance obtained by this theory is only related to the flow parameters in the cross-section of the fish's caudal. Furthermore, the theory is only applicable to analyzing the swimming of fish with small amplitude. One year later, the "large-amplitude elongated-body theory" was further proposed by Lighthill [34]. In 1991, Tong *et al.* developed the "three-dimensional waving plate theory" based on the "two-dimensional waving plate theory" of Wu [35,36]. This theory simplifies the swimming of a fish to a flexible deformed plate oscillating in a wave-like motion. It is worth noting that the tail vortex effect is considered, which makes the calculation results closer to the real swimming of the fish. This theory is applicable to fish swimming with small amplitude. It can be extended to the accelerated swimming of fish and large-amplitude non-linear swimming.

In recent years, there have been new developments in the theory of robot fishes' swimming. They are mainly a supplement to the previous theories and thus solve some practical problems. Wang *et al.* incorporated the robot fish's head oscillation equation into the kinematic model based on the elongated-body theory [33,37]. The improved kinematic model was established successfully. The results show that the maximum swing angle of the head was reduced to 86% of its original value, while the swimming speed was increased by 17%. Kirchhoff's equations of motion were utilized by Kopman *et al.* to show the dynamics of frontal link [38]. Caudal fin oscillation was modeled by Euler–Bernoulli beam theory. The influence of the fluid around the robot fish was described by the Morison equation. Finally, the dynamic equation of the robot fish propelled by soft fin was established.

## 3.2 Experimental observation

With the emergence of new experimental equipment, experimental observation has become more popular. PIV is the most effective experimental method. It is a method of measuring flow velocity that involves recording the position of particles in the flow field with multiple cameras and analyzing the images captured. The basic idea is to spread tracer particles in the flow field and then inject a pulsed laser into the measured flow field area. The images of the particles are recorded by two or more consecutive exposures. Zhu *et al.* visualized the flow field by PIV and obtained the flow field image of Tunabot during the caudal fin oscillation [24]. It is frequently necessary to construct special experimental platforms in order to meet the measurement of specific physical quantities. As shown in Figure 8, the robot fish was immersed in a tank and the swimming speed was measured [27].

## 3.3 Numerical simulation

In recent years, computer technology, computational fluid dynamics (CFD), and other disciplines have advanced rapidly. New iterations of computers have led to a significant increase in computing power, allowing some complex swimming problems to be solved. The calculation model is continuously improved in practice, resulting in increasing accuracy of the calculation. Thus, numerical simulation has made it possible to acquire accurate answers to some complex swimming problems. Currently, many research results are available. The hydrodynamic performance of fish of different shapes near the water surface using CFD was studied by Zhan *et al.* [39]. Using an incompressible Navier–Stokes flow solver based on the immersion boundary method, Liu *et al.* studied the body–fin and fin–fin interactions [40]. Han *et al.* used the same

Designs, motion mechanism, motion coordination, and communication of bionic robot fishes: a survey solver as Liu *et al.* [40,41] to simulate the swimming of the fish on the static cartesian grid. The interactions between the intermediate fins were analyzed in detail. The CFD method was used by Macias *et al.* to simulate the swimming process of the fish in undisturbed water flow [42]. Zhu *et al.* combined the immersed boundary-lattice Boltzmann method in numerical simulation with a deep recurrent Q-network to simulate the behavior of fish [43]. It provided an effective method for researching fish adaptation behaviors in complex environments. All of the above swimming problems require a massive amount of computation, which was previously extremely difficult to achieve. From the results of the calculations, all of the authors considered that the accuracy of the calculations met the requirements. We believe that numerical simulation as a method will have considerable potential in the future.

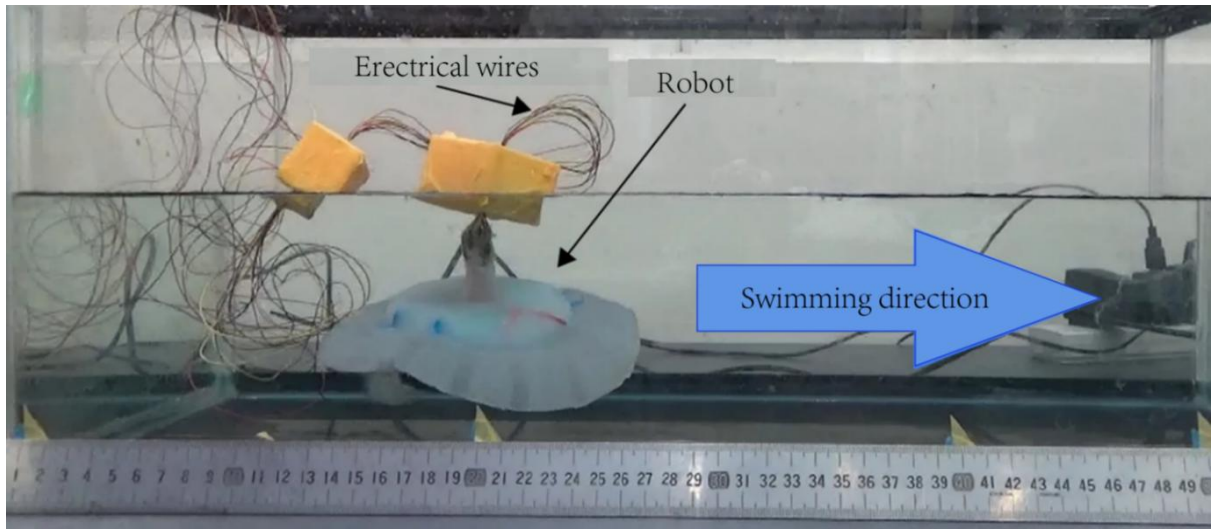

Figure 8: Experimental setup for measuring the swimming speed [27].

### 3.4 Multiple research methods

Using multiple research methods to analyze a problem, each research method can not only complement each other's strengths but also verify the results of the others, which increases the convincingness of the research. Korkmaz *et al.* established kinematic and dynamic models of the robot fish using the Denavit-Hartenberg method and Lagrange method, respectively [2]. The swimming of the robot fish was simulated using MATLAB/Simulink. Experiments in the pool validated the simulation results. Behbahani *et al.* established the dynamic model of robot fishes using the rigid body dynamics theory [44]. The hydrodynamic force acting on the pectoral fin was solved by the blade element theory. The kinetic model was evaluated experimentally. The dynamic equation of the fish in autonomous swimming was established by Xin *et al.* [45]. The steering motion of fish was simulated using three-dimensional (3D) CFD software. Liu *et al.* established a kinematic model by simplifying the caudal fin to a rigid hydrofoil and the caudal peduncle to a rigid plate [46]. The caudal fin propulsion mechanism was analyzed using CFD to determine the principle of generating propulsive power. It can be anticipated that this method will be used by more and more researchers and become a new research trend.

## 4 motion coordination and communication of multiple robot fishes

The research of multiple robot fishes emerged in recent years and is now a hot research field. When discussing the problem of multiple robot fishes, we are most concerned with the problems of motion coordination and communication of multiple robot fishes. As a result, we review the latest research on these two issues in depth.

Designs, motion mechanism, motion coordination, and communication of bionic robot fishes: a survey

## 4.1 Motion coordination of multiple robot fishes

Fish frequently congregate in schools. Fish schools can not only effectively fight against natural enemies but also save energy and help them survive in harsh environments. Researchers believe that schools of multiple robot fishes can reap the same benefits. Therefore, we focus on coordinated swimming of multiple robot fishes and related discussions. The current study is mainly concerned with tandem formation and parallel formation. However, there have been studies on other planar formations.

Tandem formation refers to the connection of the heads and tails of two or more fish in a straight line, as shown in Figure 9. The fish at the front of the line is known as the leading fish, and the fish behind it is known as the following fish. The most basic formation of this is two fish swimming in tandem formation. Tandem swimming of two 3D bionic fish was studied by Wu *et al*. [47]. The results show that, in the absence of any control by the two fish, the vortex generated by the leading fish deflected the path of the following fish. Khalid *et al*. found that the undulating frequency of the following fish does not affect the vortex and time-averaged drag of the leading fish at a certain Strouhal number [48]. Furthermore, it appeared to be more favorable for the leading fish when both fish kept swimming in tandem formation.

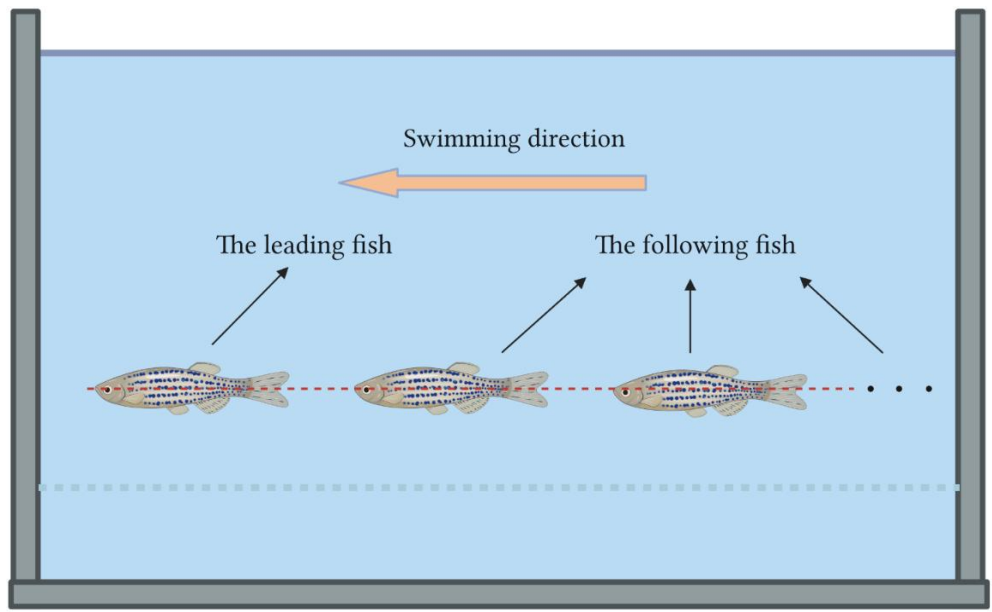

Figure 9: Tandem formation of fish.

Parallel formation refers to two or more fish lining up in a row, as shown in Figure 10. Similarly, the fish at the front of the line is called the leading fish, and the fish behind it is called the following fish. The most basic form of this is two fish swimming in parallel formation. The efficiency of two fish when swimming in parallel was analyzed by Doi *et al*. [49]. The results show that the highest swimming efficiency was achieved when the distance between the two fish (K1) was 0.4 BL under the premise of $L_1 = 0$. A vortex phase matching strategy for robot fishes was found by Li *et al*. [50]. The following robot fish could conserve energy when the front–back distance between two robot fishes was linearly connected to the tailbeat phase difference. As shown in Figure 11, the following robot fish could save energy by vortex phase matching. By fitting, the phase difference was found to be linearly related to the phase difference, as shown in Figure 11a,b. Subsequent experiments confirmed that fish also exhibit this swimming strategy. Without a complicated vision system and artificial lateral line system (ALLS), this swimming strategy can reduce energy consumption and

Designs, motion mechanism, motion coordination, and communication of bionic robot fishes: a survey enhance swimming efficiency. This is quite crucial. The swimming speed and energy consumption of a single robot fish and two parallel robot fishes were investigated by Li *et al*. [51]. It was discovered that, regardless of the tail-beat phase difference, maintaining a parallel formation always increased their swimming speed and decreased their energy consumption. Furthermore, the authors hypothesized that fish can balance their consumption with the benefits they receive from their neighbors by adjusting the tail-beat phase difference as they swim. This suggested that individuals in swimming schools might engage in competitive games.

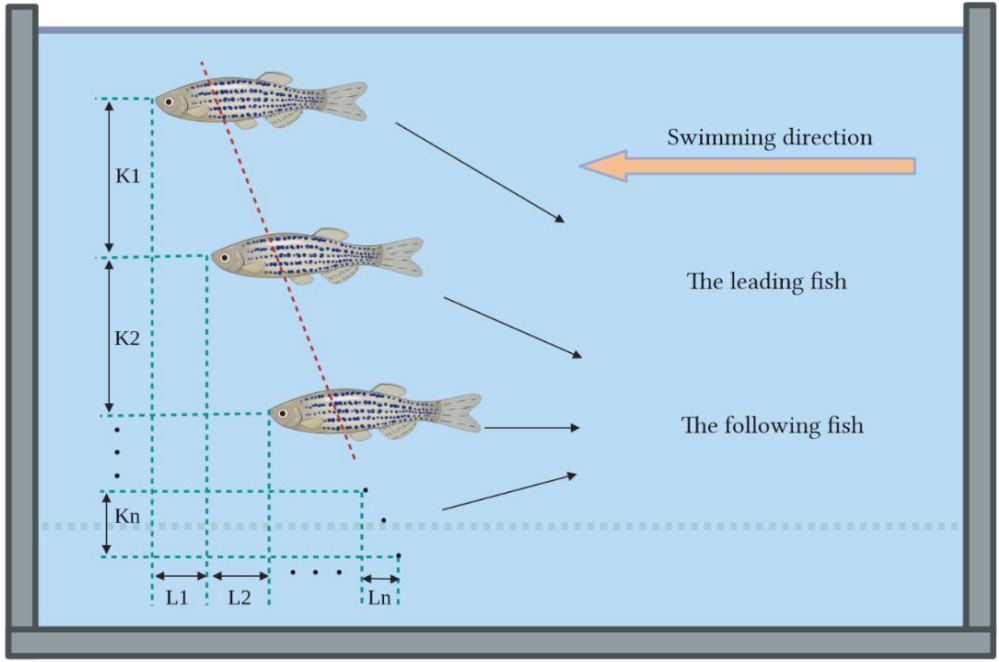

Figure 10: Parallel formation of fish ( $L_i \geq 0 \text{ and } K_i \geq 0, i = 1, 2, \cdots, n$ ).

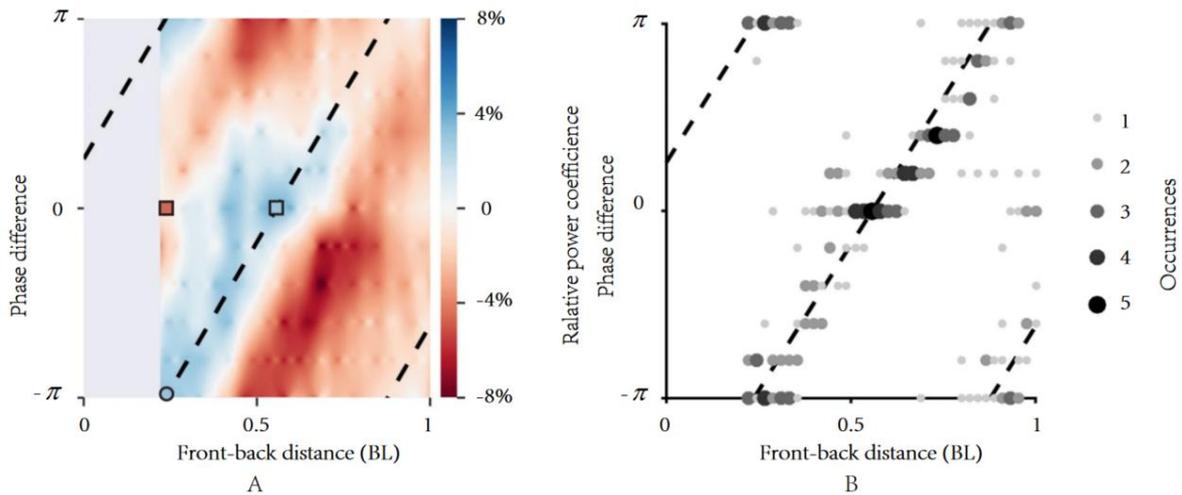

Figure 11: The following robot fish saves energy by vortex phase matching. (A) Relative power coefficient: Positive and negative values, respectively, represent energy saving and energy cost relative to swimming alone. The dashed line represents the function between phase difference and front-back distance, as shown in (B). (B) Location of energy saving: The size and darkness of the dots represent the number of times that the energy saving state occurs [50].

The discussion of various planar formations aids in determining the best formation. The average swimming efficiency of robot fish formations formed in tandem, square, diamond, and rectangular shapes was investigated by Li *et al*. [52]. It was found that the average swimming efficiency of the tandem formation was highest when the spacing of robot fishes was less than 1.25 BL. The average swimming efficiency of the rectangular formation was highest when the spacing was greater than 1.25 BL. In addition, the wake and pressure generated by the oscillation of the robot fish had an important effect on the Froude efficiency. The wake primarily influenced propulsive force, while pressure primarily influenced the lateral power loss. In this study, the phase difference of each robot fish's oscillation was constant, and the situation when the phase difference changed was not discussed.

The 3D formation is closer to a natural school of fish, and therefore it has more practical application. 3D is mainly reflected by having the height difference as a variable. The energy consumption of two robot fishes when they formed a 3D formation was studied by Li *et al*. [53]. The results show that the following robot fish could save energy consumption when there was a linear relationship between the height difference and phase difference of the two robot fishes. This research result is significant because it provided ideas for the future 3D formation of robot fishes.

**4.2 Communication of multiple robot fishes**

When multiple robot fishes form a formation, they must communicate with others in order to maintain the formation and avoid a collision. Since the distance between each robot fish is short, this is a problem for underwater close communication. Relevant studies have been conducted to date, and some solutions have been proposed. Among them, Xie Guangming's team from Peking University conducted extensive research and produced impressive results.

A proper electronic communication system facilitates the communication of multiple robot fishes. Since the robot fish's electronic communication system frequently uses the same channel, collisions always occur during communication. To solve this problem, based on carrier sense multiple access with collision avoidance (CSMA/CA), an electronic communication system was proposed by Zhang *et al*. [54]. This system incorporated a communication channel detection circuit and employed a CSMA/CA-based protocol. The simulation and experimental results validate the system's effectiveness. Nevertheless, this communication system suffered from effective bandwidth loss.

Fish can perceive information from the surrounding fluid using the lateral line system (LLS) [55]. This has serious implications for their underwater survival. Inspired by the excellent performance of the fish's LLS, the artificial lateral line system (ALLS) was designed and applied to the robot fish. Predictably, ALLS plays an important role in improving the interaction and collaboration capabilities between adjacent robot fishes. Zheng *et al*. established ALLS by composing an array of pressure sensors [56]. This ALLS could detect vortex streets generated by adjacent robot fish. According to the experimental results, it allowed the robot fish to perceive the relative vertical distance and yaw/pitch/roll angle with the adjacent robot fish. Furthermore, the oscillation amplitude/frequency/offset of the adjacent robot fish could also be sensed. However, the study was limited to the perception of the outside world by only one robot fish applying ALLS. Therefore, Zheng *et al*. further investigated ALLS on the perception of longitudinal separation sensing of two robot fishes [57]. Longitudinal separation implies that the two robot fishes maintain constant lateral spacing, change the longitudinal spacing, and keep the robot fish within the influence of the vortex produced by another robot fish. The meaning of longitudinal spacing and lateral spacing is clearly shown in Figure 12. The authors experimentally obtained a qualitative relationship between the longitudinal separation of two robot fishes and the ALLS-measured hydrodynamic pressure variations. The effectiveness of ALLS in relative state awareness applications was also verified. Unfortunately, the study was limited to qualitative analysis, with no quantitative analysis.

Using vision for communication is the most straightforward method. Berlinger *et al*. devised a new method of communication in schools of robot fishes that was inspired by the fact that fish could use vision to coordinate their

Designs, motion mechanism, motion coordination, and communication of bionic robot fishes: a survey motions [58]. The vision system of the robot fish was comprised of two cameras and LEDs. Through the algorithm, the robot fish could quickly determine the location of the adjacent robot fish after recognizing the light. The experimental results demonstrate that the robot fish could perform a variety of school behaviors using visual information. However, it is unclear whether the communication technology is still effective in environments that may hinder vision, such as murky waters.

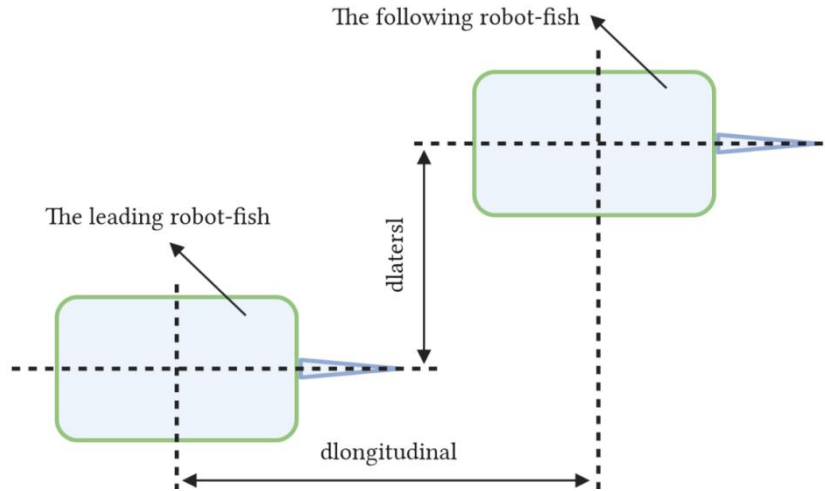

Figure 12: The meaning of longitudinal spacing and lateral spacing.

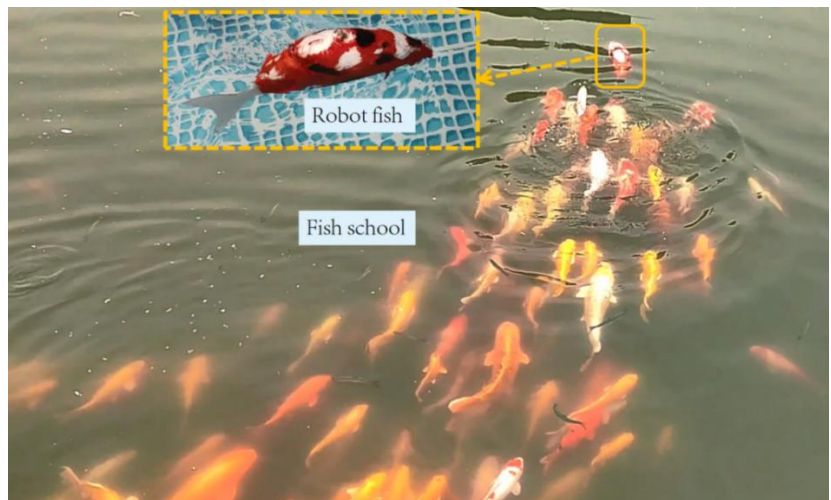

Figure 13: The robot fish and the school of fish that follows it.

We find that communication can be established between the robot fish and the fish school. When the robot fish swims in the water, it attracts the fish school to move closer to it. Eventually, the robot fish becomes the leader, leading the whole school of fish to swim forward, as shown in Figure 13. It is worth noting that the robot fish does not have smell, sound, or light to attract the fish. We hypothesize that the tail vortices created by the robot fish when it swims are the cause of this phenomenon. The swimming performance of robot fish is much inferior to that of fish. One of the reasons for this is that the tail vortices are not fully utilized. As is known, creatures have always tended to be profit-oriented. When fish perceive tail vortices, they tend to take advantage of them. In turn, it follows the robot fish, which eventually leads to this phenomenon. This brings the robot fish into communication with the fish school. We are confident that finding out how to exploit this communication would be meaningful research.

Designs, motion mechanism, motion coordination, and communication of bionic robot fishes: a survey

# 5  Challenges and future works

Thanks to a lot of research on bionic robot fishes in recent years, significant progress has been made. However, there are numerous challenges that need further work.

- Robot fishes are far inferior to fish in terms of swimming performance. Table 4 displays the performance parameters of typical robot fishes over the last five years. The maximum swimming speed of robot fishes in Table 4 is currently only 4 BL·s$^{-1}$, whereas a fish can easily reach 8 BL·s$^{-1}$ with regular swimming [59]. This demonstrates the gap in swimming performance between robot fishes and fish, which is an urgent problem to be solved. We believe there are several approaches to solve this problem. The first approach is to investigate the effect of the vortices on the swimming efficiency of robot fishes. We believe that the high propulsion efficiency of fish is closely related to the vortices they generate when they swim. It is possible to improve the swimming efficiency of robot fishes by measuring the vortices generated when fish swim and reproducing them in robot fishes. The second approach is to narrow the gap between the drive systems of robot fishes and the muscles and skin of fish. Robot fishes simulate the swimming of fish by using multiple rigid connecting rods. Fish have a flexible body made up of muscles and skin that allows them to swim continuously and supplely. However, due to the rigidity of the connecting rod and the limitation of the number of rods, the motion of robot fishes exhibits a discrete and unnatural movement. Attempts can be made to flex the connecting rod to achieve continuous motion of robot fishes, thus improving maneuverability. The third approach is to further reduce the water resistance of robot fishes when swimming. Water resistance is currently decreased mostly by designing the shape of the robot fish to be streamlined. Fish, on the other hand, have fish scales and mucous on their bodies, which can considerably reduce resistance. However, the relevant design is rarely observed in the current robot fishes. The fourth approach is to conduct an in-depth investigation of robot fishes in the BCF-MPF propulsion mode. Robot fishes in BCF propulsion mode swim fast but have poor maneuverability. In contrast, robot fishes in MPF propulsion mode have great maneuverability but slow swimming speed. The BCF-MPF propulsion mode combines the above two propulsion modes, which can accurately imitate the swimming of fish. With a reasonable design, it can achieve high swimming speed and great maneuverability and has wider application prospects. This is a promising research direction. The final approach is to use sensor technology to create close connections between robot fishes and fish. Replicating the swimming process of fish can improve the swimming performance of robot fishes. Through the sensors, we obtain real-time feedback data (body deformation, etc.) when fish swim, further completing the monitoring of the entire swimming process. Finally, the collected data are applied to robot fishes. This allows robot fishes to make rhythmic movements similar to fish, improving their swimming performance.

- The majority of studies have only used one research method to investigate the swimming mechanism of robot fishes. Actually, each research method has its own strengths and weaknesses. Because of the weaknesses, using only one method may provide unconvincing results. The combined use of multiple research methods not only achieves the complementary benefits of each method but also allows each method to verify the others to ensure the accuracy of the results.

- There is a lack of sufficient research on motion coordination and communication of multiple robot fishes. Multiple robot fishes in an appropriate formation have been shown to reduce energy consumption [47-53]. In nature, the number of fish in a school is usually greater than three, and the school is in a three-dimensional formation. However, the current study has limitations in terms of the number and formation of robot fishes.

Designs, motion mechanism, motion coordination, and communication of bionic robot fishes: a survey

Table 4: Typical robot fishes and their performance parameters.

| Reference | Maximum swimming speed | | Minimum turning radius (m) | Frequency (hertz) | | Swimming type | Structural type |
|---|---|---|---|---|---|---|---|
| | BL·s$^{-1}$ | m·s$^{-1}$ | | Caudal fin | Pectoral fins | | |
| Ref. [19] | 0.5(Ave) | 0.23(Ave) | 0.78(Ave) | NA | --- | BCF | Soft |
| Ref. [20] | 0.25 | 0.037 | NA | 0.75 | --- | BCF | Soft |
| Ref. [24] | 4 | 1.02 | NA | 15 | --- | BCF | Rigid–soft |
| Ref. [22] | 2.5 | 0.25 | NA | NA | --- | BCF | Rigid–soft |
| Ref. [25] | 0.87 | 0.31 | NA | 2.9 | --- | BCF | Rigid–soft |
| Ref. [26] | 0.45 | 0.052 | NA | --- | 1 | MPF | Soft |
| Ref. [28] | 0.94 | 0.43 | ≈ 0 | --- | NA | MPF | Rigid–soft |
| Ref. [27] | NA | 0.013 | NA | --- | 4 | MPF | Rigid–soft |
| Ref. [30] | 0.69 | 0.064 | 0.085 | NA | NA | BCF-MPF | Soft |
| Ref. [29] | NA | 0.062 | 0.234 | NA | NA | BCF-MPF | Soft |
| Ref. [32] | 0.66 | 0.365 | 0.139 | NA | NA | BCF-MPF | Rigid–soft |

NA, not available; Ave, average; BCF, body and/or caudal fin; MPF, median and/or paired fin. Frequency (hertz) indicates the value at the maximum (or Ave) swimming speed. The ranking of the references is based on the magnitude of the maximum swimming speed (BL·s$^{-1}$) of the robot fish and is classified by swimming type and structural type.

Specifically, the number of robot fishes is generally two, and the formation of robot fishes is mostly flat. The research of three or more robot fishes and the research of three-dimensional formation of robot fishes will be future research trends. The communication of multiple robot fishes is an intriguing research topic. Robot fishes need to communicate with each other to form formations, thus reducing energy consumption. The research on communication among multiple robot fishes has only recently received adequate attention. The related technology is not yet fully mature and should be tested in the actual environment. In addition, the research on communication between robot fish and fish schools is interesting content. We can imagine a future where schools of robot fishes swim together with schools of fish to form a larger school, achieving energy savings as well as harmony between robot fishes and fish.

# 6 Conclusion

This paper provides a comprehensive review of recent advances in several important fields of bionic robot fishes. The latest achievements in the development of robot fishes are presented. Based on the discussion of the main swimming theories of fish, the latest progress in the study of the swimming mechanism is summarized. The current state of research in the new field of motion coordination and communication of multiple robot fishes is analyzed.

Based on the survey, the data show that robot fishes in BCF propulsion mode can obtain high propulsion speed. This reflects the speed advantage of BCF propulsion mode. Robot fishes in MPF propulsion mode realize a small radius or even in situ turning. The turning radius is an important indicator of maneuverability. This reflects the high maneuverability of the MPF propulsion mode. The maneuverability of robot fishes in BCF-MPF propulsion mode is improved compared to robot fishes in BCF propulsion mode. However, in terms of swimming speed, compared with robot fishes in MPF propulsion mode, they fail to demonstrate the expected superiority. As a result, robot fishes in this propulsion mode have more room for advancement. The high-frequency oscillation of the caudal fin can significantly increase the propulsive speed. The soft robot fish has a low speed of propulsion compared with the rigid–soft coupled robot fish.

This paper primarily summarizes research results on robot fishes from the last five years, and the reader should be aware of the paper's time constraints. In the future, we will make efforts to improve the swimming performance of robot fishes and continue to track new advances in the research of robot fishes.

## References

[1] Triantafyllou MS, Triantafyllou GS. An efficient swimming machine. *Sci Am* 1995;272:64-70.

[2] Korkmaz D, Akpolat ZH, Soygüder S, Alli H. Dynamic simulation model of a biomimetic robotic fish with multi-joint propulsion mechanism. *Transactions of the Institute of Measurement and Control* 2015;37:684-95.

[3] Videler JJ. Fish swimming. 1st ed. London: Chapman and Hall Ltd; 1993. pp. 1-226.

[4] Lauder GV, Anderson EJ, Tangorra J, Madden PG. Fish biorobotics: kinematics and hydrodynamics of self-propulsion. *J Exp Biol* 2007;210:2767-80.

[5] Fish FE. Advantages of natural propulsive systems. *Mar Technol Soc J* 2013;47:37-44.

[6] Lauder GV. Fish locomotion: recent advances and new directions. *Ann Rev Mar Sci* 2015;7:521-45.

[7] Sfakiotakis M, Lane D, Davies J. Review of fish swimming modes for aquatic locomotion. *IEEE J Oceanic Eng* 1999;24:237-52.

[8] Breder CM. The locomotion of fishes. *Zoologica* 1926;4:159-297.

[9] Wang A. Development and analysis of body and/or caudal fin biomimetic robot fish. *J Mech Eng* 2016;52:137.

[10] Cai Y. Research advances of bionic fish propelled by oscillating paired pectoral foils. *JME* 2011;47:30.

[11] Colgate J, Lynch K. Mechanics and control of swimming: a review. *IEEE J Oceanic Eng* 2004;29:660-73.

[12] Yu J, Wen L, Ren Z. A survey on fabrication, control, and hydrodynamic function of biomimetic robotic fish. *Sci China Technol Sci* 2017;60:1365-80.

[13] Scaradozzi D, Palmieri G, Costa D, Pinelli A. BCF swimming locomotion for autonomous underwater robots: a review and a novel solution to improve control and efficiency. *Ocean Engineering* 2017;130:437-53.

[14] Yu J, Wang M, Dong H, Zhang Y, Wu Z. Motion control and motion coordination of bionic robotic fish: a review. *J Bionic Eng* 2018;15:579-98.

[15] Chu W, Lee K, Song S, et al. Review of biomimetic underwater robots using smart actuators. *Int J Precis Eng Manuf* 2012;13:1281-92.

[16] Liu H, Tang Y, Zhu Q, Xie G. Present research situations and future prospects on biomimetic robot fish. *Inter J Smart Sens Intell Syst* 2022;7:458-80.

[17] Liu G, Wang A, Wang X, Liu P. A review of artificial lateral line in sensor fabrication and bionic applications for robot fish. *Appl Bionics Biomech* 2016;2016:4732703.

[18] Raj A, Thakur A. Fish-inspired robots: design, sensing, actuation, and autonomy--a review of research. *Bioinspir Biomim* 2016;11:031001.

[19] Katzschmann RK, DelPreto J, MacCurdy R, Rus D. Exploration of underwater life with an acoustically controlled soft robotic fish. *Sci Robot* 2018;3:eaar3449.

[20] Shintake J, Cacucciolo V, Shea H, Floreano D. Soft biomimetic fish robot made of dielectric elastomer actuators. *Soft Robot* 2018;5:466-74.

[21] Liu S, Wang Y, Li Z, Jin M, Ren L, Liu C. A fluid-driven soft robotic fish inspired by fish muscle architecture. *Bioinspir Biomim* 2022;17:026009.

[22] Liao P, Zhang S, Sun D. A dual caudal-fin miniature robotic fish with an integrated oscillation and jet propulsive mechanism. *Bioinspir Biomim* 2018;13:036007.

[23] Coral W, Rossi C, Curet OM, Castro D. Design and assessment of a flexible fish robot actuated by shape memory alloys. *Bioinspir Biomim* 2018;13:056009.

[24] Zhu J, White C, Wainwright DK, Di Santo V, Lauder GV, Bart-Smith H. Tuna robotics: a high-frequency experimental platform exploring the performance space of swimming fishes. *Sci Robot* 2019;4:eaax4615.

[25] Chen B, Jiang H. Body stiffness variation of a tensegrity robotic fish using antagonistic stiffness in a kinematically singular configuration. *IEEE Trans Robot* 2021;37:1712-27.

[26] Li G, Chen X, Zhou F, et al. Self-powered soft robot in the mariana trench. *Nature* 2021;591:66-71.

[27] Yurugi M, Shimanokami M, Nagai T, Shintake J, Ikemoto Y. Cartilage structure increases swimming efficiency of underwater robots. *Sci Rep* 2021;11:11288.

[28] Ma H, Cai Y, Wang Y, Bi S, Gong Z. A biomimetic cownose ray robot fish with oscillating and chordwise twisting flexible pectoral fins. *IR* 2015;42:214-21.

[29] Zhang Z, Yang T, Zhang T, et al. Global vision-based formation control of soft robotic fish swarm. *Soft Robot* 2021;8:310-8.

[30] Li T, Li G, Liang Y, et al. Fast-moving soft electronic fish. *Sci Adv* 2017;3:e1602045.

[31] Li Z, Ge L, Xu W, Du Y. Turning characteristics of biomimetic robotic fish driven by two degrees of freedom of pectoral fins and flexible body/caudal fin. *Inter J Adv Rob Syst* 2018;15:172988141774995.

[32] Zhong Y, Li Z, Du R. Robot fish with two-DOF pectoral fins and a wire-driven caudal fin. *Advanced Robotics* 2018;32:25-36.

[33] Lighthill MJ. Aquatic animal propulsion of high hydromechanical efficiency. *J Fluid Mech* 1970;44:265.

[34] Lighthill MJ. Large-amplitude elongated-body theory of fish locomotion. *Proc R Soc Lond B* 1971;179:125-38.

[35] Tong BG, Zhuang LX. Hydrodynamic Model for Fish's Undulatory Motion and Its Applications. *Chin J Nat* 1998;1:1-7.

[36] Wu TY. Swimming of a waving plate. *J Fluid Mech* 1961;10:321-44.

[37] Wang P, Xu BZ, Lou BD, et al. Optimization and experimentation on the kinematic model of bionic robotic fish. *CAAI Trans Intell Syst* 2017;12:196-201.

[38] Kopman V, Laut J, Acquaviva F, Rizzo A, Porfiri M. Dynamic modeling of a robotic fish propelled by a compliant tail. *IEEE J Oceanic Eng* 2015;40:209-21.

[39] Zhan JM, Gong YJ, Li TZ. Gliding locomotion of manta rays, killer whales and swordfish near the water surface. *Sci Rep* 2017;7:406.

[40] Liu G, Ren Y, Dong H, Akanyeti O, Liao JC, Lauder GV. Computational analysis of vortex dynamics and performance enhancement due to body-fin and fin-fin interactions in fish-like locomotion. *J Fluid Mech* 2017;829:65-88.

[41] Han P, Lauder GV, Dong HB. Hydrodynamics of median-fin interactions in fish-like locomotion: Effects of fin shape and movement. *Physics of Fluids* 2020;32:011902.

[42] Macias MM, Souza IF, Brasil Junior AC, Oliveira TF. Three-dimensional viscous wake flow in fish swimming-A CFD study. *Mechanics Research Communications* 2020;107:103547.

[43] Zhu Y, Tian FB, Young J, Liao JC, Lai JCS. A numerical study of fish adaption behaviors in complex environments with a deep reinforcement learning and immersed boundary-lattice Boltzmann method. *Sci Rep* 2021;11:1691.

[44] Behbahani SB, Tan X. Design and modeling of flexible passive rowing joint for robotic fish pectoral fins. *IEEE Trans Robot* 2016;32:1119-32.

[45] Xin Z, Wu C. Vorticity dynamics and control of the turning locomotion of 3D bionic fish. *SAGE* 2018;232:2524-35.
[46] Liu G, Liu S, Xie Y, Leng D, Li G. The analysis of biomimetic caudal fin propulsion mechanism with CFD. *Appl Bionics Biomech* 2020;2020:7839049.
[47] Wu C, Wang L. Numerical simulations of self-propelled swimming of 3D bionic fish school. *Sci China Ser E-Technol Sci* 2009;52:658-69.
[48] Khalid MSU, Akhtar I, Dong H. Hydrodynamics of a tandem fish school with asynchronous undulation of individuals. *Journal of Fluids and Structures* 2016;66:19-35.
[49] Doi K, Takagi T, Mitsunaga Y, Torisawa S. Hydrodynamical effect of parallelly swimming fish using computational fluid dynamics method. *PLoS One* 2021;16:e0250837.
[50] Li L, Nagy M, Graving JM, Bak-Coleman J, Xie G, Couzin ID. Vortex phase matching as a strategy for schooling in robots and in fish. *Nat Commun* 2020;11:5408.
[51] Li L, Ravi S, Xie G, Couzin ID. Using a robotic platform to study the influence of relative tailbeat phase on the energetic costs of side-by-side swimming in fish. *Proc Math Phys Eng Sci* 2021;477:20200810.
[52] Li S, Li C, Xu L, Yang W, Chen X. Numerical simulation and analysis of fish-like robots swarm. *Applied Sciences* 2019;9:1652.
[53] Li L, Zheng X, Mao R, Xie G. Energy saving of schooling robotic fish in three-dimensional formations. *IEEE Robot Autom Lett* 2021;6:1694-9.
[54] Zhang H, Wang W, Zhou Y, et al. CSMA/CA-based electrocommunication system design for underwater robot groups. *IEEE/RSJ International Conference on Intelligent Robots and Systems (IROS)* 2017:2415-20.
[55] Zhai Y, Zheng X, Xie G. Fish lateral line inspired flow sensors and flow-aided control: a review. *J Bionic Eng* 2021;18:264-91.
[56] Zheng X, Wang C, Fan R, Xie G. Artificial lateral line based local sensing between two adjacent robotic fish. *Bioinspir Biomim* 2017;13:016002.
[57] Zheng XW, Wang MY, Zheng JZ, et al. Artificial lateral line based longitudinal separation sensing for two swimming robotic fish with leader-follower formation. *IEEE/RSJ International Conference on Intelligent Robots and Systems* 2019:2539-44.
[58] Berlinger F, Gauci M, Nagpal R. Implicit coordination for 3D underwater collective behaviors in a fish-inspired robot swarm. *Sci Robot* 2021;6:eabd8668.
[59] Spierts IL, Leeuwen JL. Kinematics and muscle dynamics of C- and S-starts of carp (Cyprinus carpio L.). *J Exp Biol* 1999;202:393-406.